\documentclass[10pt,twocolumn,letterpaper]{article}

\usepackage{cvpr}              

\usepackage{graphicx}
\usepackage{amsmath}
\usepackage{amssymb}
\usepackage{booktabs}

\usepackage{diagbox}  

\usepackage{multirow}
\usepackage{xcolor}
\usepackage{booktabs}
\usepackage{caption}
\usepackage[linesnumbered, vlined, ruled, commentsnumbered]{algorithm2e}
\usepackage[pagebackref,breaklinks,colorlinks]{hyperref}

\usepackage[capitalize]{cleveref}
\crefname{section}{Sec.}{Secs.}
\Crefname{section}{Section}{Sections}
\Crefname{table}{Table}{Tables}
\crefname{table}{Tab.}{Tabs.}

\def\cvprPaperID{****} 
\def\confName{CVPR}
\def\confYear{2022}

\begin{document}

\title{IDR: Self-Supervised Image Denoising via Iterative Data Refinement}

\author{
Yi Zhang\textsuperscript{1}
\qquad Dasong Li\textsuperscript{1}
\qquad Ka Lung Law\textsuperscript{2}
\qquad Xiaogang Wang\textsuperscript{1} \\
\qquad Hongwei Qin\textsuperscript{2}
\qquad Hongsheng Li\textsuperscript{1} \\ 
\textsuperscript{1}The Chinese University of Hong Kong\qquad 
\textsuperscript{2}SenseTime Research \\
{\tt\small zhangyi@link.cuhk.edu.hk}
}

\maketitle

\begin{abstract}
The lack of large-scale noisy-clean image pairs restricts supervised denoising methods' deployment in actual applications. While existing unsupervised methods are able to learn image denoising without ground-truth clean images, they either show poor performance or work under impractical settings (\eg, paired noisy images).
In this paper, we present a practical unsupervised image denoising method to achieve state-of-the-art denoising performance. 
Our method only requires single noisy images and a noise model, which is easily accessible in practical raw image denoising.
It performs two steps iteratively: (1) Constructing a noisier-noisy dataset with random noise from the noise model; (2) training a model on the noisier-noisy dataset and using the trained model to refine noisy images to obtain the targets used in the next round.
We further approximate our full iterative method with a fast algorithm for more efficient training while keeping its original high performance.
Experiments on real-world, synthetic, and correlated noise show that our proposed unsupervised denoising approach has superior performances over existing unsupervised methods and competitive performance with supervised methods.
In addition, we argue that existing denoising datasets are of low quality and contain only a small number of scenes.
To evaluate raw image denoising performance in real-world applications, we build a high-quality raw image dataset \textit{SenseNoise-500} that contains 500 real-life scenes. The dataset can serve as a strong benchmark for better evaluating raw image denoising. \textit{Code and dataset will be released\footnote{\url{https://github.com/zhangyi-3/IDR}}}
\end{abstract}


\begin{figure}[t]
   \begin{center}
      \includegraphics[width=0.99\linewidth]{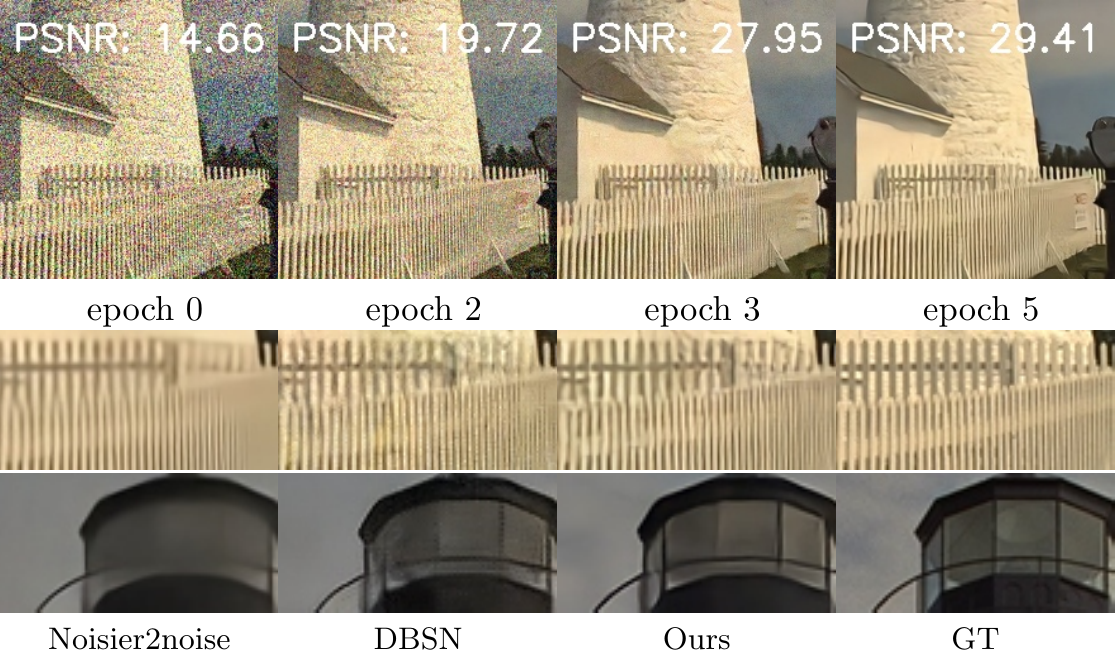}
   \end{center}
   \vspace{-8pt}
      \caption{\textbf{Top row}: The training targets created at different epochs. They have been refined progressively in our approach. 
      \textbf{Bottom row}: Comparison with previous works: DBSN \cite{DBSN} (ECCV 2020), Noisier2Noise \cite{noisier2noise} (CVPR 2020). 
      } 
   \label{fig:teaser}
\end{figure}

\section{Introduction}

In recent years, deep neural networks have made great progress in image denoising. The mainstream of the advances is powered by supervised learning-based approaches \cite{sid, cbdnet, disneyasy, kpn, unprocessing}.
However, large-scale datasets with high-quality ground-truth are still unavailable due to high human and time costs, which limit the learning-based approaches' applications in real-world scenarios.

A few unsupervised learning-based approaches have been proposed to learn image denoising without ground truth. However, existing unsupervised denoising methods either show poor performance or work under impractical settings.
One research direction focuses on unsupervised learning of denoising from single noisy images \cite{dip, n2s, n2v, DBSN,Huang_2021_CVPR}. Although such training data are convenient to collect, learning from single noisy images shows relatively poor performance since the task is highly ill-posed. 
Another research direction focuses on learning from paired noisy images \cite{n2n}. It utilizes two perfectly aligned images of the same scene as the training pairs. Such a method significantly improves the denoising performance and can sometimes even outperform supervised learning-based methods. However, multiple perfectly aligned noisy images of the same scene are sometimes challenging to capture in practice, especially for moving scenes.
One recent approach applies noise models to add synthetic noise on the already noisy images to create {\it noisier-noisy} image pairs for training a denoising network to partially capture the noise prior \cite{noisier2noise}.
However, such synthetic noisier-noisy training pairs' distributions 
cannot well fit that of the actual noisy-clean image pairs (we call such train-inference distribution mismatch the \textit{data bias} between the two datasets), which limits the method's denoising performance on actual noisy images.
There are also methods \cite{self2self, noisyAsClean} that train a separate network for denoising each image separately and show better results. However, such methods inevitably have higher computational costs.

In this paper, we propose an unsupervised denoising method that adopts a practical setting for preparing training data and achieves state-of-the-art performances. It only requires single noisy images and a noise model for training a denoising neural network. Note that the noise model is convenient to obtain in actual raw image denoising (\eg, noise profile in DNG files \cite{dng}), and can be both point-wise and correlated types (\eg, raw image noise, stripe noise, multiplicative noise).

To start with, we add noise from the known noise model to actual noisy images to create noisier-noisy dataset, which treats the created noisier images as inputs and actual noisy images as learning targets. Then, we conduct a study on learning denoising from such noisier-noisy datasets to empirically validate two findings: (1) the denoising model trained on the noisier-noisy dataset has the generalization capability of denoising actual noisy images; (2) reducing the mismatch (data bias) between the created noisier-noisy dataset and the ideal noisy-clean dataset results in a better denoising model to process the actual noisy images.

The two findings inspire us to propose the iterative data refinement (IDR) scheme, which iteratively reduces the data bias between the created noisier-noisy dataset and the ideal noisy-clean dataset to train better denoising models.
In particular, we first train a denoising model based on the noisier-noisy dataset.
According to our finding 1, the network trained on the noisier-noisy dataset can be generalized to denoise the actual noisy images.
The denoised images together with the noise model can then be used to create a new intermediate dataset for a new round of training, which regards the denoised images from the previous round as the targets.
As the data bias between the noisier-noisy and the ideal noisy-clean dataset is reduced (based on our finding 2), we can train a better denoising model on the newly constructed intermediate dataset in the next round.
This newly trained model can denoise the original noisy images again, based on which, better learning targets can be created for the next round.
This process can be performed iteratively to gradually train better denoising models. To alleviate the heavy training time cost of this iterative method, we further propose a fast version of our IDR by reducing the training epochs in each data refinement iteration.

To evaluate the denoising models' performances in real-world raw-image denoising, we collect a low-light dataset \textit{SenseNoise-500}, which contains 500 diverse scenes with perfectly aligned and high-quality ground truth.
Our method outperforms existing unsupervised methods on both real-world and synthetic noise denoising tasks with a wide range of noise types and noise levels. Other than pointwise noises, we also show our algorithm's denoising capability on typical correlated noises. All results demonstrate the effectiveness and superiority of our proposed method.

Our contributions are summarized as follows:
\begin{itemize}
   \item A self-supervised image denoising scheme, iterative data refinement (IDR), is proposed. It works under a more practical setting that only requires single noisy images and a noise model.
   
   \item We also propose a fast approximation of our full method IDR, which costs the same training time as training an ordinary denoising model while maintaining the similar performance as our full-version model.
   
   \item We collect a high-quality raw image denoising dataset for training denoising models and test their performance in real-world scenarios. It contains 500 diverse scenes, with images being well-aligned and the clean images are created with $>60$s exposure.

   \item Our method outperforms existing unsupervised methods on a wide range of noise types and noise levels. In some cases, it even shows comparable performance with fully supervised methods.
\end{itemize} 

\section{Related Work}

Image denoising has been extensively studied in the literature. In this review, we divide them into three categories: Non-learning denoising methods, learning from clean data, and learning from noisy data.
In the end, we also review some popular real-world denoising benchmarks which are widely used in other papers.\\
\noindent\textbf{Non-learning denoising methods.~}
Representative non-learning methods usually are based on different assumptions. This category of methods include anisotropic diffusion, total variation denoising \cite{rudin1992nonlinear}, wavelet domain denoising \cite{PortillaSWS03}, sparse coding \cite{MairalBPSZ09}, etc. Another important category of methods is driven by image self-similarity. Classical methods such as NLM \cite{BuadesCM05, nlm} , BM3D \cite{DabovFKE07, MakinenAF19icip} utilize similar patches to remove the noise.\\
\noindent\textbf{Learning from clean data.~}
Learning from clean data is the mainstream in image denoising. Many works \cite{sid, sidd, ntire2020} have explored learning from natural noisy and clean image pairs directly.  While the ideal training pairs can be used, the amount of those pairs is quite limited.
Another way is to use external clean images and create synthetic noisy images based on the noise model \cite{dncnn, zhang2018ffdnet, kpn, cbdnet}. As a result, there exists a natural domain gap between the real dataset and the synthetic dataset. To reduce the domain gap between synthetic images and natural noisy images, some progress \cite{unprocessing, cycleisp, CIEXYZNet} has been made to process the external data to the target domain so that the external data becomes similar to the testing data. 
Other works \cite{noiseflow, eld, CameraNoiseModeling, cbdnet, zhang2021rethinking} focus on building a more accurate noise model for the real-word noise. Generating more realistic noise for training helps the model generalize better in real applications.
\\
\noindent\textbf{Learning from noisy data.~}
One line in unsupervised learning is single-image-based \cite{dip, self2self, noisyAsClean}. They require retraining models for different images, which usually is not practical. As reported in self2self \cite{self2self}, it costs around 1.2 GPU (RTX 2080Ti) hours to retrain and test for a $256 \times 256$ image.
On the other hand, batch-based unsupervised approaches start from Noise2Noise (N2N) \cite{n2n}, which demonstrates how to learn image denoising with noisy image pairs. But noisy image pairs generally are not available in the real world and N2N also requires designing different loss functions according to the noise type.
The followed-up works try to learn image denoising by using individual noisy images. Mask-based unsupervised approaches \cite{n2s, n2v, LaineKLA19} design different mask schemas to blind pixels in the input noisy images. Then, they train the network to predict the masked pixels according to noisy pixels in the input receptive field. More recent work DBSN \cite{DBSN} uses unpaired clean images to improve unsupervised denoising results through a two-stage scheme.
Noisier2noise \cite{noisier2noise} introduces the noise model into unsupervised learning and trains the network directly for the single noise level denoising. So there still exists the dataset bias. \\
\noindent\textbf{Real-world denoising datasets.~}
Several datasets have been proposed in the literature.
Darmstadt Noise Dataset (DND) \cite{dnd}    collects 50 image pairs which can only be used for testing. And most of the images in DND have relatively low levels of noise. Smartphone Image Denoising Dataset (SIDD) \cite{sidd} was proposed to evaluate denoising on smartphones. It captures multiple times for each scene and contains 10 indoor scenes, but occasionally exhibits obvious misalignments. SID dataset \cite{sid} was captured by DSLRs with the under-exposure strategy. It uses a single long-exposure frame as the ground truth.

\section{Method}

In this section, we first introduce the creation of noisier-noisy dataset, based on which, we conduct a study on learning denoising with the noisier-noisy dataset and empirically validate two findings. We then present our self-supervised denoising framework: iterative data refinement (IDR). 
To accelerate the training process of this iterative method, we further introduce a fast version of IDR to reduce the training time while leading to similar performance as the full version of our IDR.

\subsection{Pilot Study on Data Bias of Learning Denoising}
\label{sec:pilot}

For training denoising models with a known noise model, a typical type of approaches \cite{dncnn} is to create synthetic noisy images via adding noise $n$ onto the ground-truth clean images $y$, which is denoted as $y+n$ and used as network input. We name such synthetic training data {\it noisy-clean dataset} for learning the mapping $(y+n) \rightarrow y$. However, the ground-truth clean images $y$ are generally difficult to obtain.
Another plausible way is to add noise $n$ onto the actual noisy images $x$ to create the even noisier images $x+n$, where the later noisier $x+n$ are treated as inputs and the former noisy images $x$ are used as learning targets.
We call such training data {\it noisier-noisy dataset}. Although the later dataset is much easier to obtain, there is inevitable data bias between the above two types of datasets and the denoising network trained on the later noisier-noisy dataset cannot well handle the actual noisy images $x$. 
We conduct a pilot study as follows and empirically validate two important findings.
Formally, the datasets are denoted as
\begin{equation}
   \begin{aligned}
      &\text{Noisy-clean Dataset: }\{({y}_i + {n}_i, {y}_i)\}_{i=1}^{N},\\
      &\text{Noisier-noisy Dataset: }\{({x}_i + {n}_i, {x}_i)\}_{i=1}^N, \\
      &\text{where } {x}_i = {y}_i + {n}_i.
   \end{aligned}
   \label{eq:create_dataset}
\end{equation}
Here, ${y}_i$ is a clean image (which is very difficult to obtain in practice), ${x}_i = {y}_i + {n}_i$ is its corresponding noisy image, ${n}_i$ is the random noise generated from the noise model, and ${y}_i + {n}_i$\footnote{Note that ``+'' here represents applying general noise to the images, which is {\bf not limited to only additive noise}.} denotes applying the sampled noise ${n}_i$ to the actual noisy image ${y}_i$ to create a synthetic but even noisier image.
For simplicity, we use $\{y_i+n_i, y_i\}$ and $\{x_i+n_i, x_i\}$ to denote $\{(y_i+n_i, y_i)\}_{i=0}^{N}$ and $\{(x_i+n_i, x_i)\}_{i=0}^{N}$, respectively.

\begin{figure}[t]
   \begin{center}
      \includegraphics[width=\linewidth]{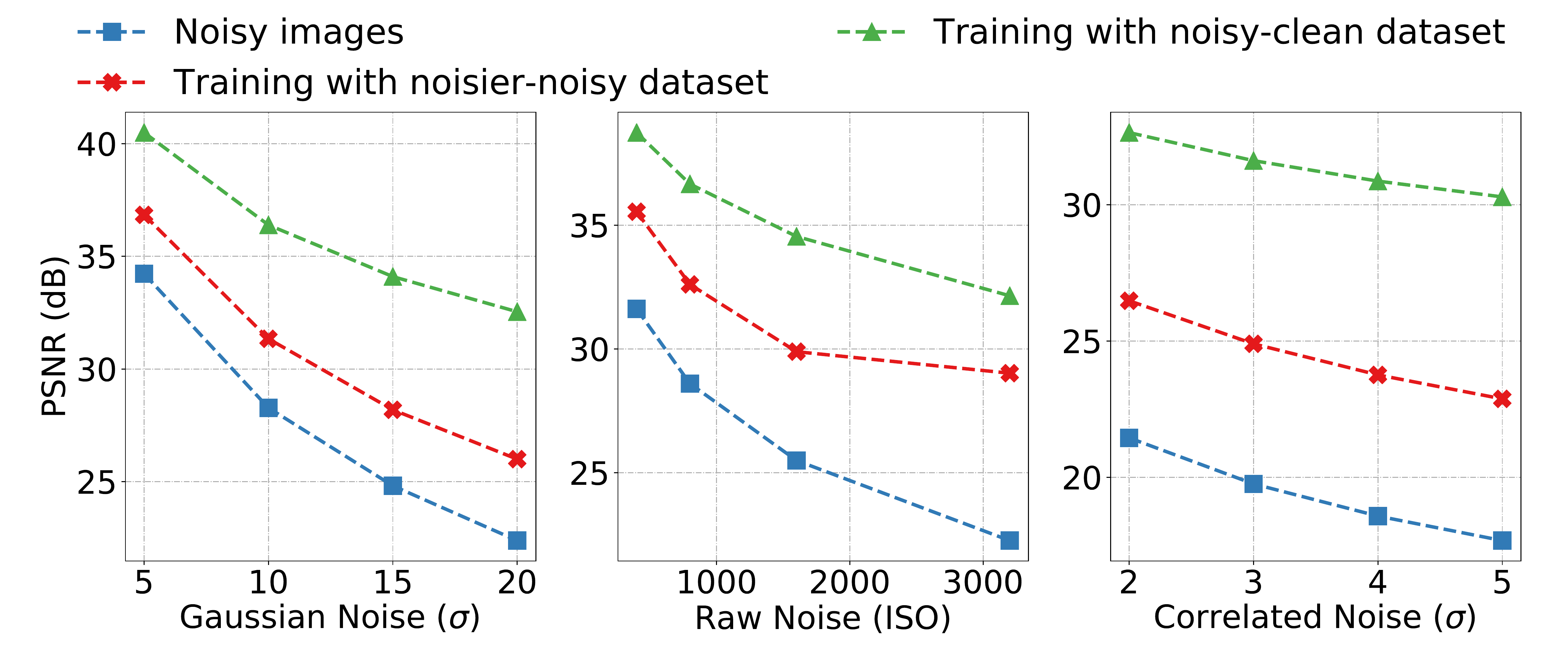}
   \end{center}
  \vspace{-8pt}
      \caption{The denoising performance of denoising networks trained with the noisy-clean dataset and the noisier-noisy datasets of three representative noises and a range of noise levels. For Gaussian noise and correlated noise, we uniformly sample their noise level from [5, 20] and [2, 5] respectively. As for the real-world raw image noise\textsuperscript{*}, we use the noise profile of a HUAWEI Mate20, which can be modeled as a Poisson-Gaussian noise model, and sample the ISO from [800, 3200] uniformly. }
      \small\textsuperscript{*}Here, the ``real-world raw image noise'' is under the assumption of the Poisson-Gaussian noise model.
   \label{fig:pilot_experiment}
\end{figure}

To analyze the data bias of training a network with noisier-noisy datasets, we train a U-Net from scratch on both the noisy-clean and noisier-noisy datasets\footnote{We randomly select 70\% images of the ImageNet validation dataset for training and the rest for testing.} and investigate 
three representative noise types, raw image, Gaussian, and correlated noise, with various noise levels. They cover common noise types \ie, signal-dependent/signal-independent noises and point-wise/correlated noises. 
More setups of the noise models can be found in the caption of Fig.~\ref{fig:pilot_experiment}.
For testing, we sample 4 discrete noise levels uniformly from the training noise levels and test the denoising models on actual noisy-clean image pairs, which have real ground truth.

As shown in Fig.~\ref{fig:pilot_experiment}, training on various noise types and noise levels show consistent results,
which lead to the following two findings:
(1) The denoising networks trained with noisier-noisy datasets can denoise the actual noisy images (red vs. blue dots in Fig.~\ref{fig:pilot_experiment}) with the same noise model.
(2) With the same noise model, training on the less biased dataset towards the ideal noisy-clean dataset helps the denoising networks achieve better denoising performance (Table \ref{tb:assumption}).

For the first finding, the models trained with noisier-noisy datasets (red dots in Fig.~\ref{fig:pilot_experiment}) always show better PSNR than the actual noisy images without any denoising (blue dots in Fig.~\ref{fig:pilot_experiment}) on all noise types, which indicate that the model can denoise actual noisy images $\{{x}_i\}$ to some extent even with the existence of data bias between the noisier-noisy and noisy-clean datasets. Recently, both single-image-based \cite{noisyAsClean} and batch-based \cite{noisier2noise} approaches have made similar conclusions on Gaussian noise but they only study a single noise level.
Our investigation extends their observations to cover a wider range of noise types and noise levels.

For the second finding, we first remove all data bias in the noisier-noisy dataset, it becomes a noisy-clean dataset (zero data bias). The model trained based on it is considered to have the optimal performances (green dots in Fig.~\ref{fig:pilot_experiment}) on all three noise types and varying noise levels.
We then synthesize different types and strengths of data bias to further verify the 2nd finding. Specifically, we apply Gaussian noise or Gaussian blur of different levels (denoted as $\sigma$ in Table \ref{tb:assumption}) to contaminate clean images $y$ to create biased targets (denoted as $y_{gn}$ and $y_{gb}$). 
The same three noise models (Gaussian, raw, correlated) with different noise levels as above are further added to the biased targets for creating noisier images ($y_{gn} + n$ and $y_{gb} + n$) as inputs for network training.
Separate denoising networks are trained with biased datasets of different strengths and are tested on actual noisy images of the same noise model. The results in Table~\ref{tb:assumption} show that the denoising networks trained on datasets with less data bias (smaller $\sigma$ in Table \ref{tb:assumption}) show better denoising performance and the second finding is consistent for two types of data bias, three different noise types, and three noise levels ($3\times 2 \times 3 =18$ separate denoising networks are trained and evaluated).

\renewcommand{\arraystretch}{1.} 
\begin{table}[t]
  \centering
  \caption{Denoising performance of models trained with biased datasets with different strengths.
  We corrupt the clean targets by applying Gaussian noise or Gaussian blur with different strengths ($\sigma$) to create biased targets ($y_{gn}$ \& $y_{gb}$) and add three different type of noise to create input noisier images ($y_{gn} + n$ \& $y_{gb} + n$). 
  The last six rows show the \textbf{PSNR drop} when models trained with different biased datasets to the ideal noisy-clean dataset (first row).
  }
  \footnotesize
  \begin{tabular}{lc|ccc} 
         \toprule
         Training    & \multirow{2}{*}{$\sigma$}   & Gaussian   & Raw  & Correlated \\
         Datasets & & Denoising & Denoising & Denoising\\
         \midrule
         $\{ {y+n, y} \}$ & - &  $30.896$  & $35.263$ & $30.484$\\
         \midrule
         \midrule
         ~ &  $5$  &  $-0.355$ & $-3.100$ & $-0.172$\\
         $\{{y}_{gn} +n , y_{gn} \}$ & $3$  & $-0.093$ & $-1.620$ & $-0.082$ \\
            
         ~ & $1$   & $-0.008$ & $-0.310$ & $-0.019$ \\
        \midrule
        \midrule
        ~ &$ 5$     & $-0.074$ & $-0.101$ & $-0.944$\\
         $\{ y_{gb} +n ,y_{gb} \}$ & $ 3$ & $-0.069$ & $-0.097$ & $-0.595$\\
         ~ &  $1$ & $-0.067$ & $-0.081$ & $-0.150$ \\
         \bottomrule
  \end{tabular}
  \label{tb:assumption}
\end{table}
\renewcommand{\arraystretch}{1}

\begin{figure*}[t]
   \begin{center}
      \includegraphics[width=0.95\linewidth]{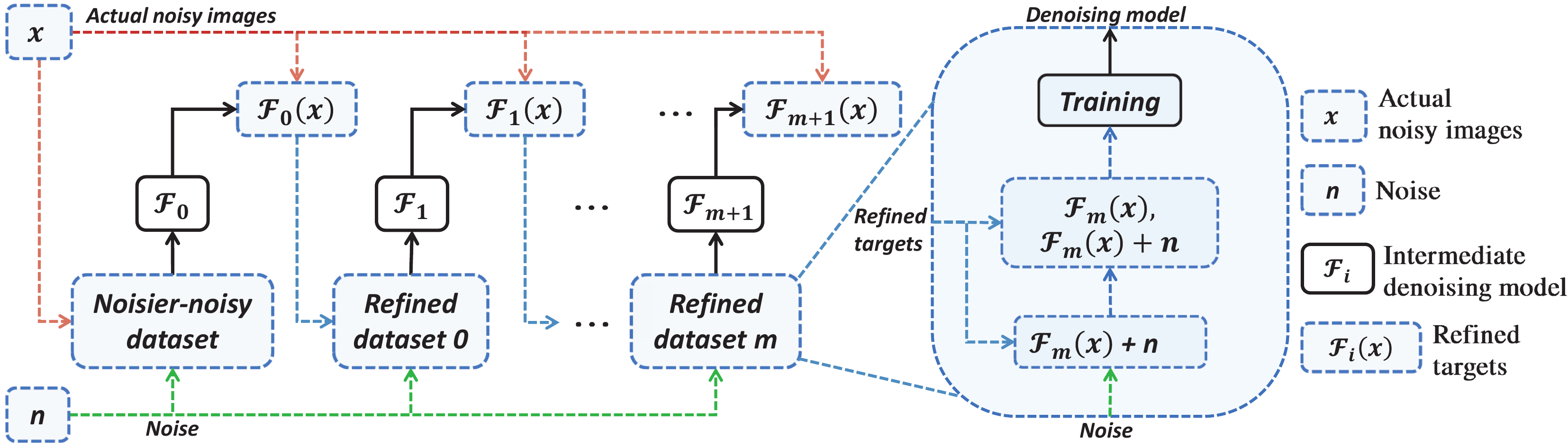}
   \end{center}
   \vspace{-10pt}
      \caption{An overview of Iterative Data Refinement (IDR).}
   \label{fig:overview}
\end{figure*}
\subsection{Iterative Data Refinement}

The two findings in Section~\ref{sec:pilot} inspire us to iteratively reduce the data bias between the noisier-noisy dataset and the ideal noisy-clean dataset. Gradually creating less biased datasets and training on them leads to promising denoising performance.

We first train a model $\mathcal{F}_{0}$ based on the initial noisier-noisy dataset $\{x_i+n_i, x_i\}$. According to our finding 1 in Section~\ref{sec:pilot}, it can denoise the actual noisy images $\{{x}_i\}$ to some extent. Therefore, we can use it to denoise the noisy images and obtain a new set of refined training dataset: 
\begin{equation}
   \{\mathcal{F}_{0}({x}_i) + {n}_i, \mathcal{F}_{0}({x}_i) \}.
\end{equation}
The newly constructed dataset shares the same number of images and scene contents as the original noisier-noisy dataset $\{{x}_i + {n}_i, {x}_i\}$. However, this new dataset has less data bias to the noisy-clean dataset since the $L_2$ distance of their targets have been reduced by our trained denoising model $\mathcal{F}_{0}$\footnote{The PSNR improvements after denoised by $\mathcal{F}_0$ indicates the $L_2$ distance to clean targets has been reduced.} (according to our finding 1 in Section \ref{sec:pilot}).

Then, using the less biased dataset construct above, we can train a new denoising model $\mathcal{F}_{1}$ from scratch in the next round:
\begin{equation}
   \mathcal{F}_{1} \leftarrow \{\mathcal{F}_{0}({x}_i) + {n}_i, \mathcal{F}_{0}({x}_i) \}.
\end{equation}
Since the data bias between the new dataset $\{\mathcal{F}_{0}({x}_i)+n_i, \mathcal{F}_{0}({x}_i)\}$ and the ideal noisy-clean dataset $\{y_i+n_i, y_i\}$ is mitigated, according to our finding 2 in Section \ref{sec:pilot}, the trained model $\mathcal{F}_{1}$ can better generalize to actual noisy images $\{{x}_i \}$ than $\mathcal{F}_0$ trained on the first-round noisier-noisy dataset.

Based on the improved model $\mathcal{F}_{1}$, the above process can be performed iteratively and alternatively by refining the training targets and training better models as shown in \cref{fig:overview}. 
From the model's perspective, once we have a denoising model $\mathcal{F}_{m}$, a better model $\mathcal{F}_{m+1}$ can be trained with the constructed dataset:
\begin{equation}
    \mathcal{F}_{m+1} \leftarrow \{\mathcal{F}_{m}({x}_i)+ {n}_i, \mathcal{F}_{m}({x}_i) \}.
\end{equation}
The previous model $\mathcal{F}_{m}$ helps to close the gap between the new noisier-noisy dataset $\{\mathcal{F}_{m}({x}_i)+n_i, \mathcal{F}_{m}({x}_i)\}$ and the ideal noisy-clean dataset $\{y_i + n_i, {y}_i\}$. In the next round of training, the improvement of the newly created noisier-noisy dataset leads to the advance of our newly trained model $\mathcal{F}_{m+1}$. 
On the other hand, from the target's perspective, a series of intermediate refined targets $\{ \mathcal{F}_{m}({x}_i) \}$ (for $m=0, 1,\dots$) are produced by the intermediate denoising models. As shown in \cref{fig:teaser}, the noise on training targets is removed and more textures are restored progressively.
Moreover, unlike traditional iterative methods that denoise one noisy image several times and lead to losing textures heavily, our method refines the training dataset and denoises the noisy image once during inference.

\begin{algorithm}[t]
   \SetAlgoLined  
   \KwIn{Noisy images $\{ {x}_i \}$; 
   Noise model ${n}$; \newline
   Total epochs $M$;  
   }
   \KwOut{Final model $\mathcal{F}_{M}$}
   \BlankLine
   Initialize the model $\mathcal{F}_0$ randomly \;
   Optimize the model $\mathcal{F}_0$ on the noisier-noisy dataset $\{{x}_i + {n}_i, {x}_i \}$ for one epoch \;
   \For{$m\leftarrow 1$ \KwTo $M$}{
      Initialize the model $\mathcal{F}_{m}  =\mathcal{F}_{m-1}$\;
      Create ``cleaner'' targets $\{{x}_i^{(m)}\} \leftarrow \{\mathcal{F}_{m-1}({x}_i)\}$\;
      Construct the new training dataset $\{{x}_i^{(m)}+ {n}_i, {x}_i^{(m)} \}$\;
      Optimize current model $\mathcal{F}_{m}$ by minimizing the $L_1$ loss for one epoch \;
   }
   \caption{Fast Iterative Data Refinement}
   \label{algo:fast_iteration} 
\end{algorithm}

\subsection{Fast Iterative Data Refinement}
\label{sec:approximation}

While the above iterative data refinement does not increase time cost in inference, its training time is proportional to the number of rounds of the proposed self-supervised data refinement.
To alleviate the heavy training time cost, we further propose a fast approximated training scheme. The whole algorithm is shown in the Algorithm \ref{algo:fast_iteration}. 

We approximate our full method in two aspects. First, we introduce one-epoch training, which performs one round of data refinement after each epoch. Specifically, for each newly constructed dataset, we use it to train our model for only one epoch instead of training it until full convergence as before. In the next epoch, we train the denoising network based on a newly constructed dataset.
We sacrifice the time needed for full model optimization but increase the iteration rounds of data refinement. The overhead we introduced is to produce refined targets before each training epoch. It costs less than 5\% of the overall training time\footnote{The forward process of producing refined targets can be performed in parallel. We use batch size 32 for the inference. For training, our batch size is 4.}. As a result, the total training time is reduced to almost the same as training a denoising model for only one round. 
 
Second, when training on the new dataset at each epoch, our models are initialized by the model from the previous epoch instead of from scratch. 
This cumulative training strategy helps the denoising network converge faster with the proposed fast data refinement scheme and ensures that the final denoising model has been optimized continuously during the whole training process. The approximation algorithm also converges fast and keeps a similar performance as our full algorithm.

\begin{figure*}[t]
   \begin{center}
      \includegraphics[width=0.95\linewidth]{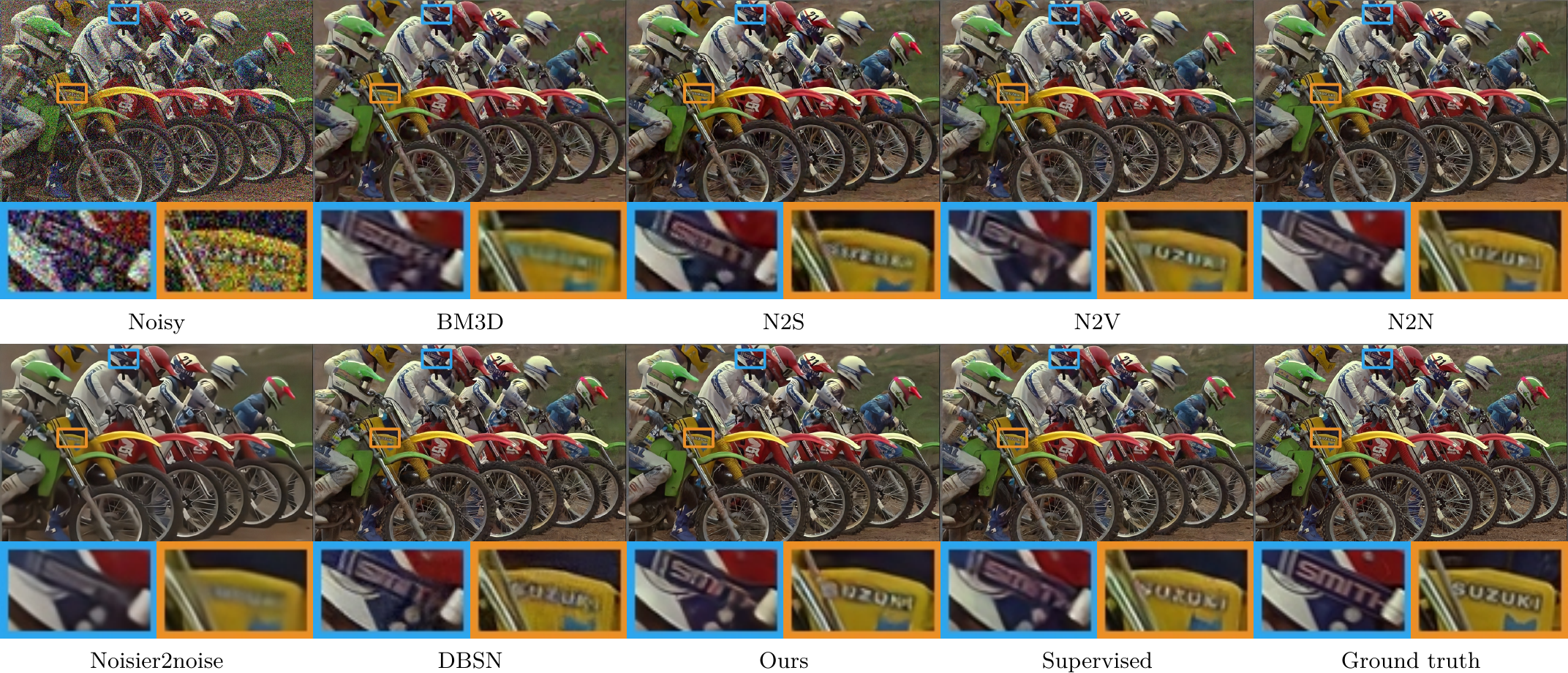}
   \end{center}
   \vspace{-10pt}
      \caption{The qualitative results of compared methods on denoising images with Gaussian noise ($\sigma=50$).}
   \label{fig:gaussian_noise}
\end{figure*}

\section{SenseNoise-500 dataset}\label{sec:dataset}

There exists an obvious gap between existing raw image denoising benchmarks and real-world scenarios. For example, SID \cite{sid} was created with under-exposure photos which are rare in daily photography. The mobile benchmark SIDD \cite{sidd} captures multiple images for each scene (10 indoor scenes). And, it suffers from global misalignments occasionally (e.g., see image Nos. 101 \& 52 in SIDD) due to the different autofocus or OIS.
To evaluate the denoising performance in real-world scenarios, we propose a new high-quality dataset under the common low-light conditions using a popular mobile CMOS sensor: IMX586 with $3000 \times  4000$ pixels.  
The dataset contains 500 scenes and is larger than the commonly used denoising benchmarks: SIDD (10 scenes) \cite{sidd}, DND dataset (50 scenes) \cite{dnd} and SID Sony (231 scenes) \cite{sid}.
It was collected across 3 cities and covers common indoor and outdoor scenes in daily life.  
When we construct our dataset, there are some important considerations:

\noindent\textbf{High-quality ground truth.~} The ground truths in the existing denoising benchmarks~\cite{sid, sidd} are still more or less noisy, especially at high noise levels. To obtain high-quality ground truths, we capture a sequence of 64 frames for each scene, among which, 4 frames of them are captured in normal exposure with the default camera setting and used as noisy testing images. 
The other 60 frames are long-exposure for creating the ground truth. For each pixel location, we collect all pixel values at the same location in the 60 long-exposure images and use the median value as the ground truth.
The exposure time for each long-exposure frame lies in $[1s, 2s]$ , which means that the total exposure time for creating each ground truth is between 60s and 120s. It guarantees that all ground truths are noise-free even at the highest noise level.

\noindent\textbf{Alignment and noise calibration.~}
To obtain well-aligned noisy/clean image pairs, we capture each image sequence on the tripod and control the mobile camera with a wireless shutter. All camera parameters are set and fixed in advance through camera2 API \cite{camera2api}. In this way, both normal-exposure and long-exposure frames are captured successively without any movement to the device. On the other hand, we try to avoid large object motions and dynamic illuminations in both indoor and outdoor scenes.  Under this setting, all frames (long-exposure and normal-exposure) captured for one scene are perfectly aligned.
While some companies may provide the noise profiles in raw files (\eg, DNG \cite{dng}), they are usually not accurate. We re-calibrate the noise parameters of IMX586 under the noise formulation in ELD \cite{eld}. However, we choose the Gaussian distribution to model the read noise, since we observe that it is more robust for mobile sensors. More details and examples of \textit{SenseNoise-500} can be found in the supplementary materials.

\renewcommand{\arraystretch}{1.1} 

\begin{table*}[!t]
  \centering 
  \caption{The PSNR/SSIM results of Gaussian denoising on both sRGB datasets (Kodak and BSDS300) and the gray dataset (BSD64). 
  One superior supervised method (MPRnet\cite{Zamir2021MPRNet}) is also provided for comparison.
  Best and second best results are \textbf{highlighted} and \underline{underlined}. }
  \footnotesize
  \begin{tabular}{lc|ccccccc|c}
         \toprule
         Dataset & $\sigma$  & BM3D \cite{MakinenAF19icip} & N2V \cite{n2v}    & Nr2n \cite{noisier2noise} & DBSN \cite{DBSN} & N2N \cite{n2n} & Supervised  & Ours & MPRnet~\cite{Zamir2021MPRNet} \\ 
         \midrule
         \multirow{2}{*}{Kodak}  & 25  & 31.88 / 0.869&	31.63 / 0.869 & 31.96 / 0.869 &32.07 / 0.875&\textbf{32.39} / \textbf{0.886}&	\underline{32.39} / \underline{0.885}&	32.36 / 0.884&	32.85 / 0.893
         \\ 
         ~ & 50 & 	28.64 / 0.772&	28.57 / 0.776 &28.73 / 0.770&	28.81 / 0.783	&29.23 / 0.803&	\underline{29.24} / \underline{0.803}&	\textbf{29.27} / \textbf{0.803}&	29.78 / 0.817
         \\
         \hline   
         BSDS & 25 & 	30.47 / 0.863&	30.72 / 0.874	 &	29.57 / 0.815&31.12 / 0.881&	31.39 / 0.889&	\underline{31.40} / \underline{0.889}&	\textbf{31.48} / \textbf{0.890}&	31.87 / 0.897
         \\ 
         ~300  & 50 	&27.14 / 0.745	&27.60 / 0.775 & 26.18 / 0.684&	27.87 / 0.782&	\underline{28.17} / \underline{0.799}&	\underline{28.17} / \underline{0.799}&	\textbf{28.25} / \textbf{0.802}&	28.65 / 0.814
         \\ \hline
         \multirow{2}{*}{BSD64}  & 25 & 28.55 / 0.782&	27.64 / 0.781  & NA &	28.81 / 0.818&	29.15 / 0.831&	\underline{29.20} / \underline{0.835}&	\textbf{29.20} / \textbf{0.835} &	29.40 / 0.840
         \\
         ~ & 50 & 25.59 / 0.670&	25.46 / 0.681	 	&NA&25.95 / 0.703&	26.23 / 0.725	&\underline{26.24} / \textbf{0.727}&	\textbf{26.25} / \underline{0.726} &	26.48 / 0.736
         \\ \hline
         Average & ~  &	28.71 / 0.784&	28.60 / 0.793  & 29.11 / 0.785 & 29.11 / 0.807&	29.43 / 0.822&	\underline{29.44} / \underline{0.823}&	\textbf{29.47} / \textbf{0.823} &	29.84 / 0.833
         \\
         \bottomrule
  \end{tabular}
  \label{tb:gaussian}
\end{table*}
\renewcommand{\arraystretch}{1.} 
\section{Experiments}

To evaluate our method, we follow previous works to start with synthetic noises on popular datasets. Then, we move to tackle more challenging real-world raw image denoising, including both DSLR and mobile datasets. Moreover, we conduct ablation studies to show the generalization of our method and make further comparisons. \textbf{More experiments can be found in supplementary materials.}

\renewcommand{\arraystretch}{1.2} 
\begin{table*}[t]
   \centering
   \caption{Denoising performances (PSNR/SSIM) of different compared methods on raw image denoising. For the SID dataset \cite{sid}, we follow ELD \cite{eld} to show the results on ISO 100, 250, and 300. For our SenseNoise-500 dataset, we divide it into normal scenes (ISO $\leq 12800$) and extreme scenes (ISO $> 12800$). Best and second best results are \textbf{highlighted} and \underline{underlined}.}
  \footnotesize
   \begin{tabular}{lc|ccccc|ccc} 
         \toprule
         Dataset & ISO &  BM3D \cite{MakinenAF19icip} & N2S \cite{n2s} & N2V \cite{n2v}  & N2N \cite{n2n} &  Ours & Supervised & MIT5K+unprocess \\ 
         \midrule
         \multirow{3}{*}{SID}  & 100 & 32.92 / 0.758&	35.61 / 0.830&	35.64 / 0.827&	37.83 / 0.898	&	\underline{38.02 / 0.906} &\textbf{38.60 / 0.912} & 37.67 / 0.884
         \\ 
         ~ & 250 &29.56 / 0.686	&31.39 / 0.698	&30.58 / 0.655	&33.44 / 0.814&	\underline{33.88 / 0.819} 	&\textbf{37.08 / 0.886} & 33.14 / 0.767
         \\
         ~ & 300 & 28.88 / 0.674	&29.78 / 0.649	&28.88 / 0.595&	31.61 / 0.769	&	\underline{32.14 / 0.779}&\textbf{36.29 / 0.874}  & 31.36 / 0.727
         \\ 
         \hline   
         SenseNoise-500 & $\leq 12800$ & 41.05 / 0.958 &31.05 / 0.867 &	41.86 / 0.961&	41.22 / 0.956&	\underline{44.01 / 0.978}&	\textbf{44.80 / 0.981} & 43.36 / 0.978 \\ 
         ~  & $> 12800$ & 37.54 / 0.940 &27.95 / 0.871 &	38.28 / 0.948	&39.53 / 0.965&	\underline{40.17 / 0.966}&	\textbf{40.23 / 0.966} & 39.92 / 0.965 \\
         \bottomrule
   \end{tabular}
   \label{tb:raw_denoising}
\end{table*}
\renewcommand{\arraystretch}{1.} 
\subsection{Implementation details}
\noindent\textbf{Training details.~}
In all experiments, we use our fast IDR for comparisons. The network architecture and training settings are kept the same as N2N \cite{n2n}. Specifically, we use a shallow U-Net architecture without batch normalization layers as our backbone. All models are trained from scratch with batch size 4 for 50k iterations. We use the Adam optimizer and $L_1$ loss function.  The initial learning rate is fixed to $3\times 10^{-4}$, and then halves twice at iteration 25k and 40k.  Its input and output channels are differently set to adapt to different denoising tasks (\eg, gray-image denoising has 1 input channel).

\noindent\textbf{Datasets for Synthetic Experiments.~}
For synthetic noises, we randomly extract patches of size $256 \times 256$ from the ImageNet validation dataset for training. 
For testing datasets, we select three largest of them for evaluation: Kodak (24 images) and BSDS300 (300 images) for color images, and BSD68 (68 images) for gray-scale images.
Following the previous work \cite{n2n}, we test on four diverse noise types: (1) Gaussian noise with training on the continuous noise levels $ \sigma \in [0, 50]$ and testing on $\sigma = 25 \text{ and } 50$, (2) binomial and impulse noise with training on $p \in [0, 0.95]$ and testing on $p = 0.5$, and (3) correlated noise with $\sigma=1$.

\noindent\textbf{Datasets for Real-world Experiments.~}
For real-world raw image denoising, we train and test on the SID dataset (DSLR) \cite{sid} and the proposed \textit{SenseNoise-500} dataset (smartphones). Those two datasets cover different camera sensors and low-light conditions.
The training and testing sets for SID dataset are the same as SID \cite{sid}. For \emph{SenseNoise-500} dataset, we randomly split 400 images for training and the other 100 images for testing.
The patch size has been raised to $512 \times 512$ for better handling the low-light condition. As for the noise model, we recalibrate the noise profile using the recent ELD noise model \cite{eld} for both SID dataset and our \textit{SenseNoise-500} dataset.

\subsection{Main results}
Due to the space limitation, we only put the results of Gaussian noise and real-world noise in the main text. Please find the results of other noises in supplementary materials.

\noindent\textbf{Details of Gaussian noise experiments.~}
We compare our method with six learnining-based methods (N2N \cite{n2n}, N2V \cite{n2v}, N2S \cite{n2s}, DBSN \cite{DBSN} and Noisier2noise \cite{noisier2noise}.) and also the representative traditional method BM3D \cite{MakinenAF19icip}. For BM3D, we use the authors' refined implementation\footnote{https://pypi.org/project/bm3d/}.
For learning-based methods, all of them use the same U-Net architecture except for DBSN \cite{DBSN} provided by the authors, which is $10 \times$ larger (97.4G FLOPS) than U-Net (8.2G FLOPS). For DBSN, we use the author's released models. All other methods are retrained using the author's codes and keep the same training settings as N2N.

\noindent\textbf{Results of Gaussian noise experiments.~}
The PSNR and SSIM results are shown in Table \ref{tb:gaussian}. Our method outperforms all unsupervised methods on both color and gray images. And, surprisingly, our method even surpasses the supervised method slightly in most cases. 
Both our method and N2N introduce new training targets to unsupervised denoising, and  show clear gains compared with other methods that only use single noisy images.
But, the well-aligned paired noisy images required by N2N are difficult to capture in practice. On the contrary, the noise model used in our method can be easily calibrated and already saved in raw images. They can be easily obtained or recalibrated \cite{eld,zhang2021rethinking} in read-world raw image denoising. Some qualitative results are shown in Fig. \ref{fig:gaussian_noise}. Our method restores more textures and colors while keeping the high-quality denoising results. \\

\begin{figure}[t]
   \begin{center}
      \includegraphics[width=0.95\linewidth]{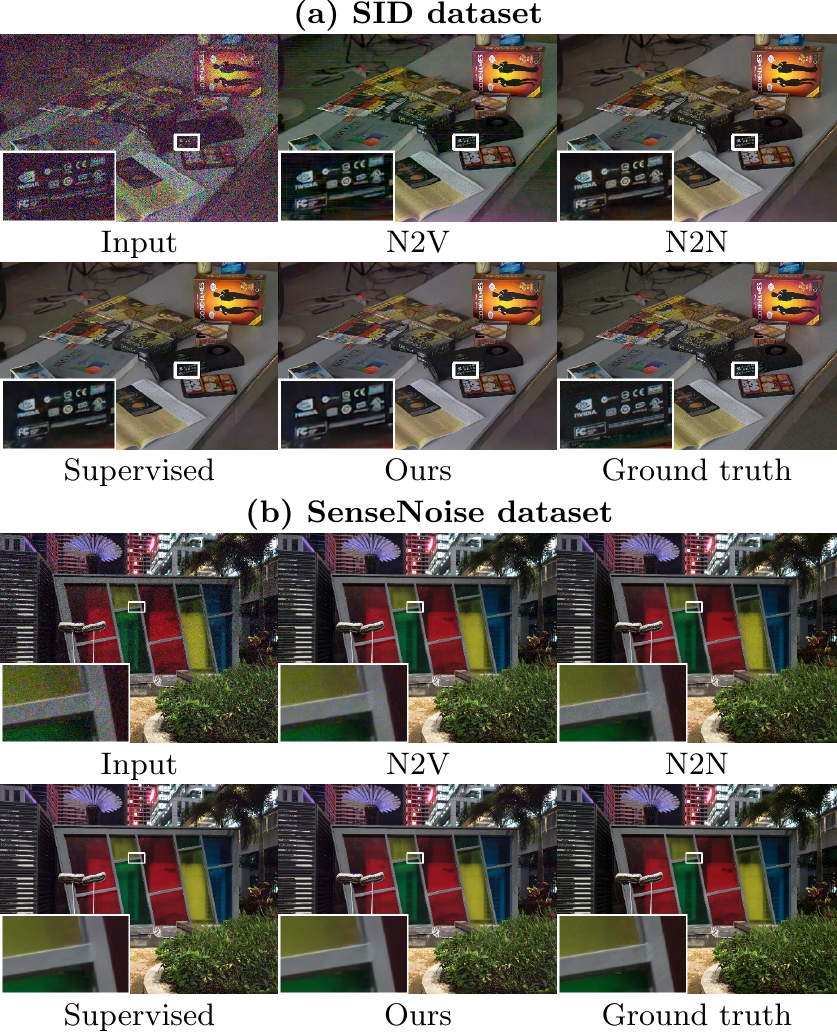}
   \end{center}
   \vspace{-10pt}
      \caption{Qualitative results of raw image denoising on SID and \textit{SenseNoise-500} datasets. (\textbf{Zoom in for details})} 
   \label{fig:raw_noise}
\end{figure}

\noindent\textbf{Details of real-world raw noise experiments.~}
For BM3D, we use the mixture Gaussian noise model to estimate the noise level. For learning-based methods, we re-train all methods by adopting the authors' code.

\noindent\textbf{Results of real-world raw noise experiments.~}
The results are shown in Table \ref{tb:raw_denoising}. Some qualitative results are shown in Fig. \ref{fig:raw_noise}. Our method performs consistently better than other unsupervised methods on both DSLR and smartphones datasets.
Both the traditional method BM3D \cite{MakinenAF19icip} and the unsupervised methods N2V \cite{n2v} and N2S \cite{n2s} with single noisy images cannot work well and show more than 2 dB performance drop compared with our method. 
Since the real-world noises are more complicated, they may not satisfy the zero-mean assumption in N2N \cite{n2n}. As a result, N2N with paired noisy images shows 0.4 dB performance drop compared with our method. 

There exists an obvious performance gap between unsupervised and supervised learning methods, especially at higher noise levels. We observe that the gap derives from two aspects. First, in extremely ill-posed conditions, some information might never be recovered without seeing the clean targets. The higher the ISO, the more obvious this phenomenon is. The second aspect might be the inaccurate noise profile. The noise distributions might be slightly different between the DSLRs of the same camera model (\eg fix-pattern noise).  

\begin{table}[t]
   \centering

   \caption{Comparing our method with supervised learning on raw image denoising. The supervised model is trained on the unprocessed \cite{unprocessing} external clean sRGB data (MIT5K \cite{fivek} and SID sRGB).}
  \footnotesize
   \begin{tabular}{lc|ccc} 
         \toprule
         Dataset  & ISO & MIT5K sRGB & SID sRGB  & Ours \\ 
         \midrule
         \multirow{3}{*}{SID} 
          & 100 & 37.67 / 0.884 & 37.76 / 0.887 & 38.02 / 0.906 \\ 
         ~& 250 & 33.14 / 0.767 & 33.24 / 0.777 & 33.88 / 0.819 \\
         ~& 300 & 31.36 / 0.727 & 31.50 / 0.741 & 32.14 / 0.779 \\
         \bottomrule
   \end{tabular}
   \label{tb:raw_unprocessing_comparision}
\end{table}
\section{Comparison with supervised learning from external clean sRGB images.~}
Once knowing the noise model, we can synthesize data for supervised training. So, an interesting comparison of our method would be the supervised training with the synthetic data, especially for the real-world raw image denoising. 
But, the large-scale clean raw image dataset for each sensor is hard and expensive to collect. Following the common practice, we unprocess the external clean sRGB image to obtain clean raw images~\cite{unprocessing}.

We conduct the study on the SID raw dataset. For the supervised method, we collect the clean sRGB images from MIT5K dataset~\cite{fivek} and unprocess them to obtain raw images \cite{unprocessing} with proper parameters. For a fair comparison, we keep the same amount of data and the noise model for both methods. The results are shown in \cref{tb:raw_unprocessing_comparision}. Our methods could still outperform supervised learning with external clean ground-truths due to the dataset content bias. Then, we remove the image content bias by unprocessing the SID sRGB images for the supervised method, its performance improves over the training on MIT5K, but the performance gap with our method still exists as shown in \cref{tb:raw_unprocessing_comparision}. 
While the supervised method (``clean image+simulated noise'') uses the same noise model as ours, our method outperforms it over 0.5dB (averaged PSNR 34.17dB vs. 34.68dB) on SID raw dataset. 
The reason is that there still exists a clear gap (image content, raw image characters) between the synthetic and real clean raw images (unprocessed MIT5K/SID sRGB vs. SID raw).  

\section{Limitations and conclusions}
In this paper, we propose a practical and iterative denoising method (IDR) that outperforms the existing unsupervised denoising methods. Our method is based on empirical findings on data bias of learning denoising. While lacking theoretical support, experiments on three synthetic noises, real-world raw image noise, and even spatially-correlated noise demonstrate the consistent robustness and effectiveness of our method.
Moreover, we build a high-quality low-light denoising dataset for evaluating the denoising performance in real-world scenarios. The dataset contains 500 diverse scenes, each with high-quality ground truth. We expect the dataset can serve as a real-world benchmark for image denoising. 
 
\section{Acknowledgement}
This work is supported in part by Centre for Perceptual and Interactive Intelligence Limited, in part by the General Research Fund through the Research Grants Council of Hong Kong under Grants (Nos. 14204021, 14207319, 14203118, 14208619), in part by Research Impact Fund Grant No. R5001-18, in part by CUHK Strategic Fund.

{\small
\bibliographystyle{ieee_fullname}
\bibliography{egbib}

\begin{thebibliography}{10}\itemsep=-1pt

\bibitem{ntire2020}
Abdelrahman Abdelhamed, Mahmoud Afifi, Radu Timofte, Michael~S. Brown, et~al.
\newblock {NTIRE} 2020 challenge on real image denoising: Dataset, methods and
  results.
\newblock In {\em {CVPR} Workshops}, pages 2077--2088. {IEEE}, 2020.

\bibitem{noiseflow}
Abdelrahman Abdelhamed, Marcus~A Brubaker, and Michael~S Brown.
\newblock {Noise Flow: Noise Modeling with Conditional Normalizing Flows}.
\newblock In {\em International Conference on Computer Vision (ICCV)}, 2019.

\bibitem{sidd}
Abdelrahman Abdelhamed, Stephen Lin, and Michael~S. Brown.
\newblock A high-quality denoising dataset for smartphone cameras.
\newblock In {\em IEEE Conference on Computer Vision and Pattern Recognition
  (CVPR)}, June 2018.

\bibitem{CIEXYZNet}
Mahmoud Afifi, Abdelrahman Abdelhamed, Abdullah Abuolaim, Abhijith
  Punnappurath, and Michael~S Brown.
\newblock Cie xyz net: Unprocessing images for low-level computer vision tasks.
\newblock {\em arXiv preprint}, 2020.

\bibitem{n2s}
Joshua Batson and Lo{\"{\i}}c Royer.
\newblock Noise2self: Blind denoising by self-supervision.
\newblock In {\em {ICML}}, volume~97 of {\em Proceedings of Machine Learning
  Research}, pages 524--533. {PMLR}, 2019.

\bibitem{unprocessing}
Tim Brooks, Ben Mildenhall, Tianfan Xue, Jiawen Chen, Dillon Sharlet, and
  Jonathan~T Barron.
\newblock Unprocessing images for learned raw denoising.
\newblock In {\em IEEE Conference on Computer Vision and Pattern Recognition
  (CVPR)}, 2019.

\bibitem{BuadesCM05}
Antoni Buades, Bartomeu Coll, and Jean{-}Michel Morel.
\newblock A non-local algorithm for image denoising.
\newblock In {\em {CVPR} {(2)}}, pages 60--65. {IEEE} Computer Society, 2005.

\bibitem{nlm}
Antoni Buades, Bartomeu Coll, and Jean{-}Michel Morel.
\newblock Non-local means denoising.
\newblock {\em Image Process. Line}, 1, 2011.

\bibitem{fivek}
Vladimir Bychkovsky, Sylvain Paris, Eric Chan, and Fr{\'e}do Durand.
\newblock Learning photographic global tonal adjustment with a database of
  input / output image pairs.
\newblock In {\em The Twenty-Fourth IEEE Conference on Computer Vision and
  Pattern Recognition}, 2011.

\bibitem{CameraNoiseModeling}
Ke-Chi Chang, Ren Wang, Hung-Jin Lin, Yu-Lun Liu, Chia-Ping Chen, Yu-Lin Chang,
  and Hwann-Tzong Chen.
\newblock Learning camera-aware noise models.
\newblock In {\em European Conference on Computer Vision (ECCV)}, 2020.

\bibitem{sid}
Chen Chen, Qifeng Chen, Jia Xu, and Vladlen Koltun.
\newblock Learning to see in the dark.
\newblock In {\em The IEEE Conference on Computer Vision and Pattern
  Recognition (CVPR)}, June 2018.

\bibitem{DabovFKE07}
Kostadin Dabov, Alessandro Foi, Vladimir Katkovnik, and Karen~O. Egiazarian.
\newblock Image denoising by sparse 3-d transform-domain collaborative
  filtering.
\newblock {\em {IEEE} Trans. Image Processing}, 16(8):2080--2095, 2007.

\bibitem{scn}
Yuchen Fan et~al.
\newblock Scale-wise convolution for image restoration.
\newblock In {\em {AAAI}}, 2020.

\bibitem{cbdnet}
Shi Guo, Zifei Yan, Kai Zhang, Wangmeng Zuo, and Lei Zhang.
\newblock Toward convolutional blind denoising of real photographs.
\newblock {\em 2019 IEEE Conference on Computer Vision and Pattern Recognition
  (CVPR)}, 2019.

\bibitem{Huang_2021_CVPR}
Tao Huang, Songjiang Li, Xu Jia, Huchuan Lu, and Jianzhuang Liu.
\newblock Neighbor2neighbor: Self-supervised denoising from single noisy
  images.
\newblock In {\em Proceedings of the IEEE/CVF Conference on Computer Vision and
  Pattern Recognition (CVPR)}, pages 14781--14790, June 2021.

\bibitem{dng}
ADOBE~SYSTEMS INCORPORATED.
\newblock Digital negative (dng) specification.
\newblock
  \url{https://www.adobe.com/content/dam/acom/en/products/photoshop/pdfs/dng_spec_1.4.0.0.pdf},
  2012.

\bibitem{n2v}
Alexander Krull, Tim-Oliver Buchholz, and Florian Jug.
\newblock Noise2void-learning denoising from single noisy images.
\newblock In {\em Proceedings of the IEEE Conference on Computer Vision and
  Pattern Recognition}, pages 2129--2137, 2019.

\bibitem{LaineKLA19}
Samuli Laine, Tero Karras, Jaakko Lehtinen, and Timo Aila.
\newblock High-quality self-supervised deep image denoising.
\newblock In {\em NeurIPS}, pages 6968--6978, 2019.

\bibitem{n2n}
Jaakko Lehtinen, Jacob Munkberg, Jon Hasselgren, Samuli Laine, Tero Karras,
  Miika Aittala, and Timo Aila.
\newblock Noise2noise: Learning image restoration without clean data.
\newblock In {\em {ICML}}, volume~80 of {\em Proceedings of Machine Learning
  Research}, pages 2971--2980. {PMLR}, 2018.

\bibitem{liu2021Swin}
Ze Liu, Yutong Lin, Yue Cao, Han Hu, Yixuan Wei, Zheng Zhang, Stephen Lin, and
  Baining Guo.
\newblock Swin transformer: Hierarchical vision transformer using shifted
  windows.
\newblock {\em arXiv preprint arXiv:2103.14030}, 2021.

\bibitem{camera2api}
Google LLC.
\newblock Android camera2 api.
\newblock
  \url{http://developer.android.com/reference/android/hardware/camera2/package-summary.html},
  2016.

\bibitem{MairalBPSZ09}
Julien Mairal, Francis~R. Bach, Jean Ponce, Guillermo Sapiro, and Andrew
  Zisserman.
\newblock Non-local sparse models for image restoration.
\newblock In {\em {ICCV}}, pages 2272--2279. {IEEE} Computer Society, 2009.

\bibitem{MakinenAF19icip}
Ymir M{\"{a}}kinen, Lucio Azzari, and Alessandro Foi.
\newblock Exact transform-domain noise variance for collaborative filtering of
  stationary correlated noise.
\newblock In {\em {ICIP}}, pages 185--189. {IEEE}, 2019.

\bibitem{kpn}
Ben Mildenhall, Jonathan~T Barron, Jiawen Chen, Dillon Sharlet, Ren Ng, and
  Robert Carroll.
\newblock Burst denoising with kernel prediction networks.
\newblock In {\em IEEE Conference on Computer Vision and Pattern Recognition
  (CVPR)}, 2018.

\bibitem{noisier2noise}
Nick Moran, Dan Schmidt, Yu Zhong, and Patrick Coady.
\newblock Noisier2noise: Learning to denoise from unpaired noisy data.
\newblock In {\em {CVPR}}, pages 12061--12069. {IEEE}, 2020.

\bibitem{dnd}
Tobias Plotz and Stefan Roth.
\newblock Benchmarking denoising algorithms with real photographs.
\newblock In {\em {CVPR}}, pages 2750--2759. {IEEE} Computer Society, 2017.

\bibitem{PortillaSWS03}
Javier Portilla, Vasily Strela, Martin~J. Wainwright, and Eero~P. Simoncelli.
\newblock Image denoising using scale mixtures of gaussians in the wavelet
  domain.
\newblock {\em {IEEE} Trans. Image Processing}, 12(11):1338--1351, 2003.

\bibitem{self2self}
Yuhui Quan, Mingqin Chen, Tongyao Pang, and Hui Ji.
\newblock Self2self with dropout: Learning self-supervised denoising from
  single image.
\newblock In {\em {CVPR}}, pages 1887--1895. {IEEE}, 2020.

\bibitem{rudin1992nonlinear}
Leonid~I Rudin, Stanley Osher, and Emad Fatemi.
\newblock Nonlinear total variation based noise removal algorithms.
\newblock {\em Physica D: nonlinear phenomena}, 60(1-4):259--268, 1992.

\bibitem{dip}
Dmitry Ulyanov, Andrea Vedaldi, and Victor~S. Lempitsky.
\newblock Deep image prior.
\newblock In {\em {CVPR}}, pages 9446--9454. {IEEE} Computer Society, 2018.

\bibitem{disneyasy}
Thijs Vogels, Fabrice Rousselle, Brian McWilliams, Gerhard R{\"{o}}thlin, Alex
  Harvill, David Adler, Mark Meyer, and Jan Nov{\'{a}}k.
\newblock Denoising with kernel prediction and asymmetric loss functions.
\newblock {\em {ACM} Trans. Graph.}, 37(4):124:1--124:15, 2018.

\bibitem{eld}
Kaixuan Wei, Ying Fu, Jiaolong Yang, and Hua Huang.
\newblock A physics-based noise formation model for extreme low-light raw
  denoising.
\newblock In {\em IEEE Conference on Computer Vision and Pattern Recognition},
  2020.

\bibitem{DBSN}
Xiaohe Wu, Ming Liu, Yue Cao, Dongwei Ren, and Wangmeng Zuo.
\newblock Unpaired learning of deep image denoising.
\newblock In {\em European Conference on Computer Vision (ECCV)}, 2020.

\bibitem{noisyAsClean}
Jun Xu, Yuan Huang, Li Liu, Fan Zhu, Xingsong Hou, and Ling Shao.
\newblock Noisy-as-clean: Learning unsupervised denoising from the corrupted
  image.
\newblock {\em CoRR}, abs/1906.06878, 2019.

\bibitem{cycleisp}
Syed~Waqas Zamir, Aditya Arora, Salman Khan, Munawar Hayat, Fahad~Shahbaz Khan,
  Ming-Hsuan Yang, and Ling Shao.
\newblock Cycleisp: Real image restoration via improved data synthesis.
\newblock In {\em CVPR}, 2020.

\bibitem{Zamir2021MPRNet}
Syed~Waqas Zamir et~al.
\newblock Multi-stage progressive image restoration.
\newblock In {\em CVPR}, 2021.

\bibitem{dncnn}
Kai Zhang, Wangmeng Zuo, Yunjin Chen, Deyu Meng, and Lei Zhang.
\newblock Beyond a {Gaussian} denoiser: Residual learning of deep {CNN} for
  image denoising.
\newblock {\em IEEE Transactions on Image Processing}, 26(7):3142--3155, 2017.

\bibitem{zhang2018ffdnet}
Kai Zhang, Wangmeng Zuo, and Lei Zhang.
\newblock Ffdnet: Toward a fast and flexible solution for {CNN} based image
  denoising.
\newblock {\em IEEE Transactions on Image Processing}, 2018.

\bibitem{zhang2021rethinking}
Yi Zhang, Hongwei Qin, Xiaogang Wang, and Hongsheng Li.
\newblock Rethinking noise synthesis and modeling in raw denoising.
\newblock In {\em Proceedings of the IEEE/CVF International Conference on
  Computer Vision (ICCV)}, pages 4593--4601, October 2021.

\end{thebibliography}
}

\newpage
\appendix

\renewcommand{\arraystretch}{1.2}
\begin{table}[t]
   \centering
   \caption{Denoising performance (PSNR/SSIM) on binomial (\textbf{B}) and impulse noise (\textbf{I}). The experiment on Kodak and BSDS300 dataset shows consistent results. Best and second best results are \textbf{highlighted} and \underline{underlined}. }
  \footnotesize
   \begin{tabular}{lc|ccc} 
         \toprule
         Dataset  & Noise & N2N \cite{n2n} & Supervised  & Ours \\ 
         \midrule
         \multirow{2}{*}{Kodak} 
          & \textbf{B} & 31.48 / 0.939 & \textbf{31.84 / 0.946} & \underline{31.63 / 0.944} \\ 
         ~& \textbf{I} & \textbf{36.04 / 0.975} & \underline{35.83 / 0.978} & 34.92 / 0.977 \\ \hline
         \multirow{2}{*}{BSDS300} 
           & \textbf{B} & 31.50 / 0.930 & \textbf{32.32 / 0.938} & \underline{32.18 / 0.936} \\ 	
         ~ & \textbf{I} & 36.55 / 0.976 & \textbf{37.48 / 0.980} & \underline{37.40 / 0.979} \\
         \bottomrule
   \end{tabular}
   \label{tb:binomial}
\end{table}
\renewcommand{\arraystretch}{1.}

\section{Experiments on other synthetic noises}
\begin{figure*}[b]
   \begin{center}
      \includegraphics[width=0.95\linewidth]{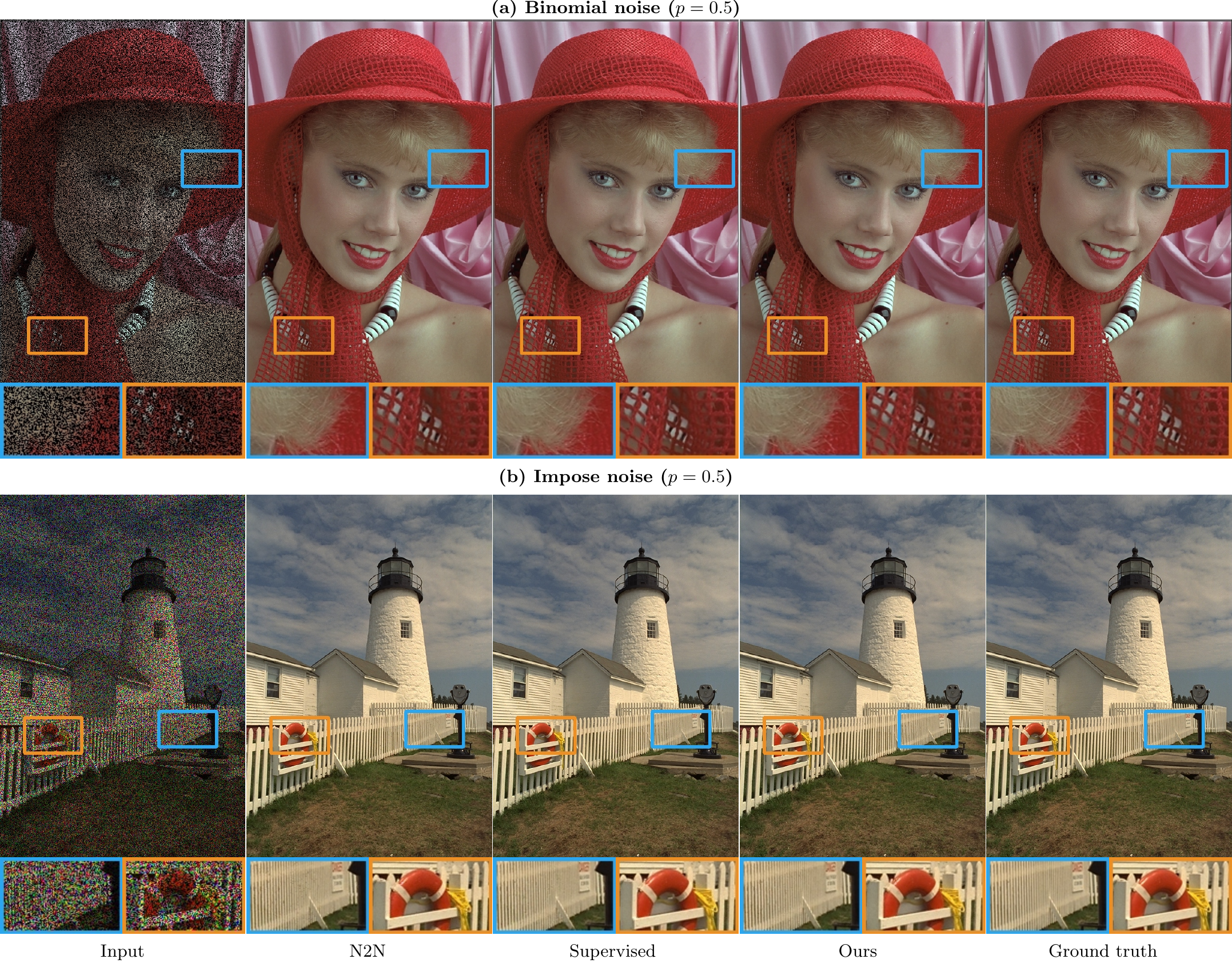}
      \end{center}
         \caption{Visualization of removing binomial and impulse noise with the corrupted probability $p=0.5$.
          While we only use individual noisy images for training, our qualitative results are very closed to supervised learning.
          } 
      \label{fig:ber_noise}
\end{figure*}

\noindent\textbf{Binomial and impulse noise.~}
We experiment with another two point-wise noise types, binomial noise and impulse noise following the setup of N2N \cite{n2n}. They can be used to model bad pixels and hot pixels in raw images, respectively. Binomial noise is constructed from the pixel-wise production of a clean image with a random 0-1 mask. It uses a one-channel mask to set some pixels to zeros with probability $p$ and retain other pixels' values. 
For impulse noise, it randomly sets some pixel channels to 0 or 1 with probability $p$ while keeping the other colors. Some examples are shown in Fig. \ref{fig:ber_noise}.
During training, we uniformly set $p \in [0, 0.95]$ for both binomial noise and impulse noise to train models on a range of noise levels. During inference, we follow the setting of N2N \cite{n2n} and fix $p = 0.5$. 
Since they are not zero-mean noises, we have to use different losses for training as recommended in N2N.

The results are shown in Table Fig. \ref{fig:ber_noise}.
Similar to the results on Gaussian noise, our method still outperforms N2N in most cases. N2N produces better results on binomial denoising in the Kodak dataset, which shows 0.2 dB improvement than supervised learning. But it cannot show stable performance when testing on the larger BSDS300 dataset.

\begin{table}[t]
   \centering
   \caption{The comparison between our method and the refined BM3D \cite{MakinenAF19icip} on correlated noise. Best results are \textbf{highlighted}.
   (See the text for more details.)}
  \footnotesize
   \begin{tabular}{l|ccc} 
         \toprule
         Dataset & Kernel & BM3D \cite{MakinenAF19icip} & Ours \\ 
         \midrule
          \multirow{5}{*}{Kodak} 
            & $g_1$ & 31.56 / 0.900 & \textbf{37.46 / 0.976} \\ 
          ~ & $g_2$ & 29.20 / 0.798 & \textbf{33.10 / 0.929} \\ 	
          ~ & $g_3$ & 31.38 / 0.857 & \textbf{41.88 / 0.993} \\
          ~ & $g_4$ & 27.23 / 0.742 & \textbf{28.60 / 0.803} \\ 
          ~ & $g_5$ & 26.53 / 0.859 & \textbf{32.97 / 0.973} \\ \hline
         \multirow{5}{*}{BSDS300} 
           & ${g}1$ & 26.77 / 0.841 & \textbf{29.83 / 0.922} \\ 
         ~ & ${g}2$ & 27.84 / 0.787 & \textbf{31.28 / 0.910} \\
         ~ & ${g}3$ &	29.59 / 0.838 & \textbf{38.22 / 0.982} \\
         ~ & ${g}4$ &	26.14 / 0.722 & \textbf{27.39 / 0.793} \\
         ~ & ${g}5$ & 26.31 / 0.870 & \textbf{33.69 / 0.974} \\
         \bottomrule
   \end{tabular}
   \label{tb:correlated_denoising}
\end{table}

\noindent\textbf{Correlated noise.~}
Most previous methods assume that the noise is pixel-wise independent so that the spatially correlated noise is less explored. Our method only requires the noise model no matter whether it is pixel-wise independent or not. We compare our method with the recently refined BM3D \cite{MakinenAF19icip} on several typical correlated noises. 
The spatially correlated noise can be created as follows:
\begin{equation}
  {x} = {y} + {v} \otimes {g} , \quad {v} \sim \mathcal{N}(0,1),
\end{equation}
where ${x}$ and ${y}$ denote the noisy and clean images, ${v}$ is the random noise generated from a normal distribution with zero mean and $\sigma=1$, and ${g}$ is the convolution kernel for creating the correlated noise. Following the refined BM3D \cite{MakinenAF19icip}, we use 5 types of kernels for comparison.

Some qualitative results are shown in Fig. \ref{fig:correlated_noise}. More results and kernel visualization are provided in the supplementary materials. We also use BSDS300 as our sRGB benchmark and evaluate the default noise level $\sigma=5$ for all correlated noise. The refined BM3D is noise-aware and our models are trained on $\sigma=5$.
In Table \ref{tb:correlated_denoising}, we show the quantitative results on correlated noise. Our methods can deal with spatially correlated noise and outperform the refined BM3D by a large margin.

\section{More ablation studies}
\renewcommand{\arraystretch}{1.1} 
\begin{table}[t]
   \centering
   \caption{The Gaussian denoising results (PSNR/SSIM) of our method when using different network architectures. All models are trained with the same training settings.}
    \small
   \begin{tabular}{l|cc|l} 
         \toprule
          Models & $\sigma=25$   & $\sigma=50$ & FLOPS(G)  \\ 
         \midrule
          U-Net &  31.84 / 0.875 & 28.71 / 0.787 &  6.0\\ 
          SCN~\cite{scn} & 32.23 / 0.879 & 29.11 / 0.795 & 54.5 \\
          MPRNet~\cite{Zamir2021MPRNet} & 32.59 / 0.887 & 29.51 / 0.808 & 300.6 \\
          \midrule
          Swin~\cite{liu2021Swin} & 32.22 / 0.878 & 29.04 / 0.792 & 16.0
          \\
         \bottomrule
   \end{tabular}
   \label{tb:diff_architecture}
\end{table}
\renewcommand{\arraystretch}{1.} 
\renewcommand{\arraystretch}{1.2} 

\begin{table}
\centering
\caption{The \textbf{training time} (hours) and \textbf{PSNR} comparison between the full version and fast version iterative data refinement. ``IDR-n" denotes the testing results after the n-th iteration in the full version iterative data refinement (IDR). ``Fast" denotes fast IDR.}
\small
\begin{tabular}{l|ccccc|c} 
\toprule
Model & IDR-1 & IDR-2 & IDR-3 & IDR-4 & IDR-5 & Fast   \\ \hline
PSNR  & 24.84 & 28.01 & 31.35 & 31.48 & 31.49 & 31.50  \\
Time & 8.3 & 16.6 & 24.9  & 33.2  & 41.5  & 8.6    \\
\bottomrule
\end{tabular}
\label{tb:hours}
\end{table}
\renewcommand{\arraystretch}{1.} 

\noindent\textbf{The generality to Transformer and other CNNs.~}In the above experiments, we follow N2N \cite{n2n} and only use U-Net as our backbone. 
Here, to show the generality of our method, we adopt both Swin Transformer~\cite{liu2021Swin} and recent CNN-based architectures (SCN \cite{scn} and MPRNet \cite{Zamir2021MPRNet}) in our iterative algorithm. 
To reduce the training time, we train those heavy models for 20k iterations and other settings are the same as our Gaussian denoising experiment.
As shown in Table~\ref{tb:diff_architecture}, using larger CNN-based denoising models in our IDR can always produce better results. Swin Transformer shows comparable results as the SCN but with much fewer flops.
This indicates our method can well adapt to both CNN and transformer architectures, and it can be further improved by advanced architectures.

\noindent\textbf{Comparison between IDR and fast IDR.~}
In Table~\ref{tb:hours}, we show the intermediate testing results and the corresponding training time cost of our full version IDR on Gaussian denoising. After about 5 full iterations, the full version method converges and shows quite similar performances as the fast IDR. But, due to the one-epoch training and the accumulative training strategy of the fast IDR, the training time is quite close to the full version algorithm.

\section{More details of SenseNoise-500 dataset}
\begin{figure*}[b]
   \begin{center}
      \includegraphics[width=0.7\linewidth]{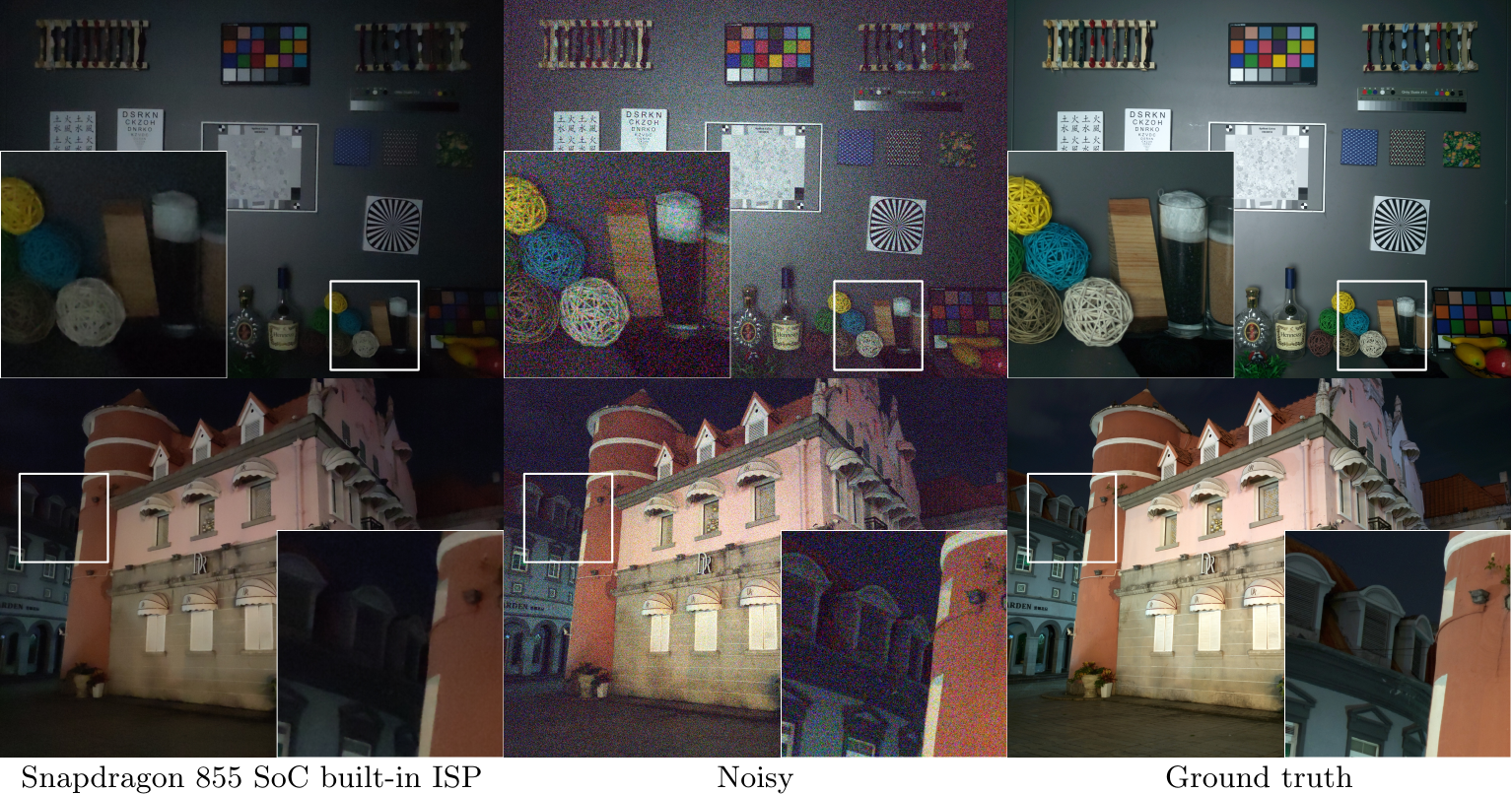}
      \end{center}
         \caption{Examples from the SenseNoise-500 dataset. Even in the extreme low-light indoor (ISO 25600) and outdoor (ISO 24379) scenes, our ground truths are still of high-quality. 
          } 
      \label{fig:ds_example}
\end{figure*}

To ensure that the collected pairs have the same brightness levels, the long-exposure and normal-exposure frames are captured with the same target sensitivity value. Specifically, for the noisy images (normal-exposure), all of them use the default exposure time and ISO recommended by the smartphone (Xiaomi MI9). And, the long-exposed frames are captured with the exposure time $t_l$ starting from 1s and up to 2s, and they keep the same target sensitivity value $S$ via a dynamic ISO ${\rm ISO}_l$:
\begin{equation}
   \begin{aligned}
      t_l &= \max\left(\min\left( \frac{S}{100}, 2\right), 1\right),~~{\rm ISO}_l = \frac{S}{t_l},
   \end{aligned} 
\end{equation}
where $S = {\rm ISO}_r \times t_r$ is the fixed target sensitivity value, ${\rm ISO}_r$ and $t_r$ are recommended ISO and exposure time for noisy images, and $\frac{S}{100}$ is the estimated maximum exposure time and it would be clipped to be within $[1s, 2s]$. To demonstrate the quality of our dataset, we show the extreme indoor and outdoor examples in Fig. \ref{fig:ds_example}.

\textbf{Object motion:} We carefully avoid capturing dynamic scenes and light changing scenes and remove unsatisfied scenes when we construct the dataset. 
As for occasional slight object motions (pixel-level), they may cause some abnormal pixel values across the temporal dimension around the motion areas. 
But those outliers can be removed naturally since we use the median filter along the temporal dimension to create the ground truth.

\textbf{OIS:} We used Xiaomi Mi 9 with the IMX586 sensor to collect the dataset. The lens optical stabilization mode can be disabled via camera2 API, which is particularly helpful to capture long-exposure images.

\textbf{The noise levels and diversity:} We write an App to show the real-time exposure information on the screen. This helps us to cover a wide range of ISOs. 
We follow the advice to compare the SNR and the standard deviation (STD) of SNR.
SenseNoise-500 SNR range: $[-12.94, 5.79]$ STD: $11.54$. SIDD SNR range: $[-12.72, 4.57]$ STD: $11.05$.
This indicates that our dataset covers a wider range and keeps better diversity compared with the SIDD dataset. 

\textbf{GT quality:} 
Since the scene contents and exposure strategy vary greatly for different datasets, we are unable to compare the SNR of their ground truth directly.
The GT for most noisy scenes (ISO25600) can be founded in \cref{fig:ds_example}.

\section{Performance gain from large-scale unlabelled data}
We add an experiment to compare supervised baseline (UNet and following Table 4's settings) using limited data with the unsupervised training and large-scale unlabeled data on Gaussian noise (\cref{tb:data_scale}).
All data are from ImageNet.
The data size for supervised methods is set to 0.5k since existing denoising datasets generally contain hundreds of images. 
When using the same training scale (0.5k), our method produces slightly lower performance than the supervised method on Gaussian noise. 
Scaling up the unlabeled data size (50k) helps to improve the performance by $\sim$0.2dB over supervised training with limited data.

\begin{table}[t]
   \centering
   \caption{Training with unlabelled data of different scales.}
   \footnotesize
   \begin{tabular}{c|cc|ccc} 
         \toprule
          ~ & \multicolumn{2}{c|}{Supervised} & \multicolumn{3}{c}{Unsupervised (Ours)} \\ \hline
          \diagbox{Sigma}{Data scale} & 0.3k  & 0.5k  & 0.5k & 5k & 50k \\ 
         \midrule
         $\sigma=25$ & 31.60 & 31.70 & 31.65 & 31.82 & 31.84 \\ 
         $\sigma=50$ & 28.53 & 28.60 & 28.61 & 28.70 & 28.71 \\
         \bottomrule
   \end{tabular}
   \label{tb:data_scale}
\end{table}

\begin{figure*}[t]
   \begin{center}
      \includegraphics[width=0.95\linewidth]{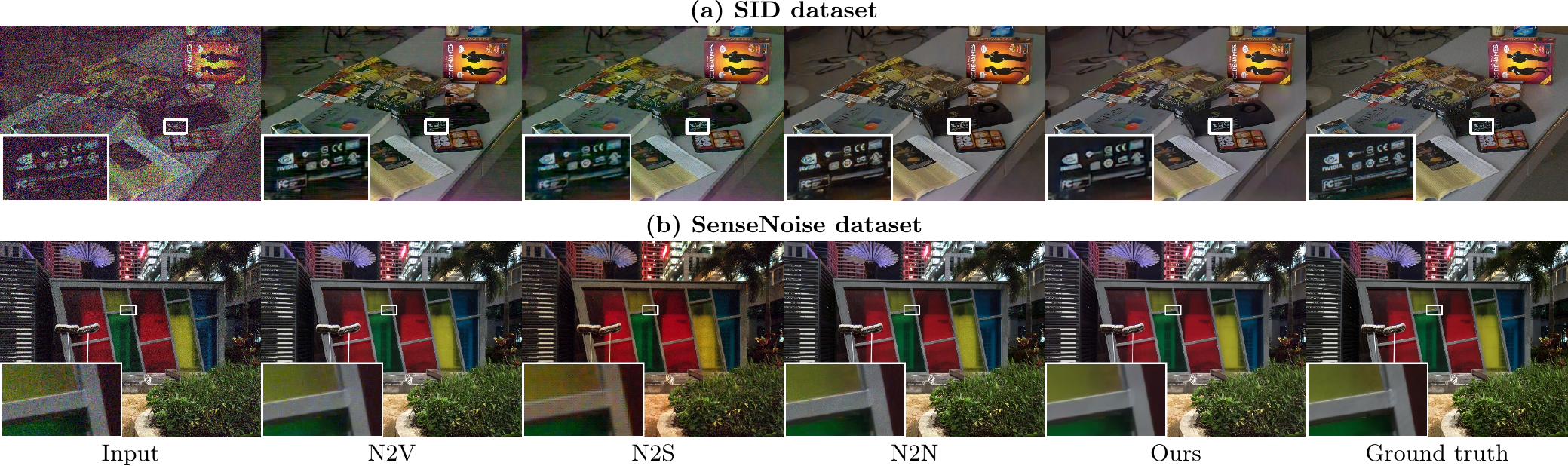}
   \end{center}
      \caption{Qualitative results of raw image denoising on SID dataset and our SenseNoise-500 dataset.} 
   \label{fig:raw_noise_f}
\end{figure*}

\begin{figure*}[t]
   \begin{center}
      \includegraphics[width=0.95\linewidth]{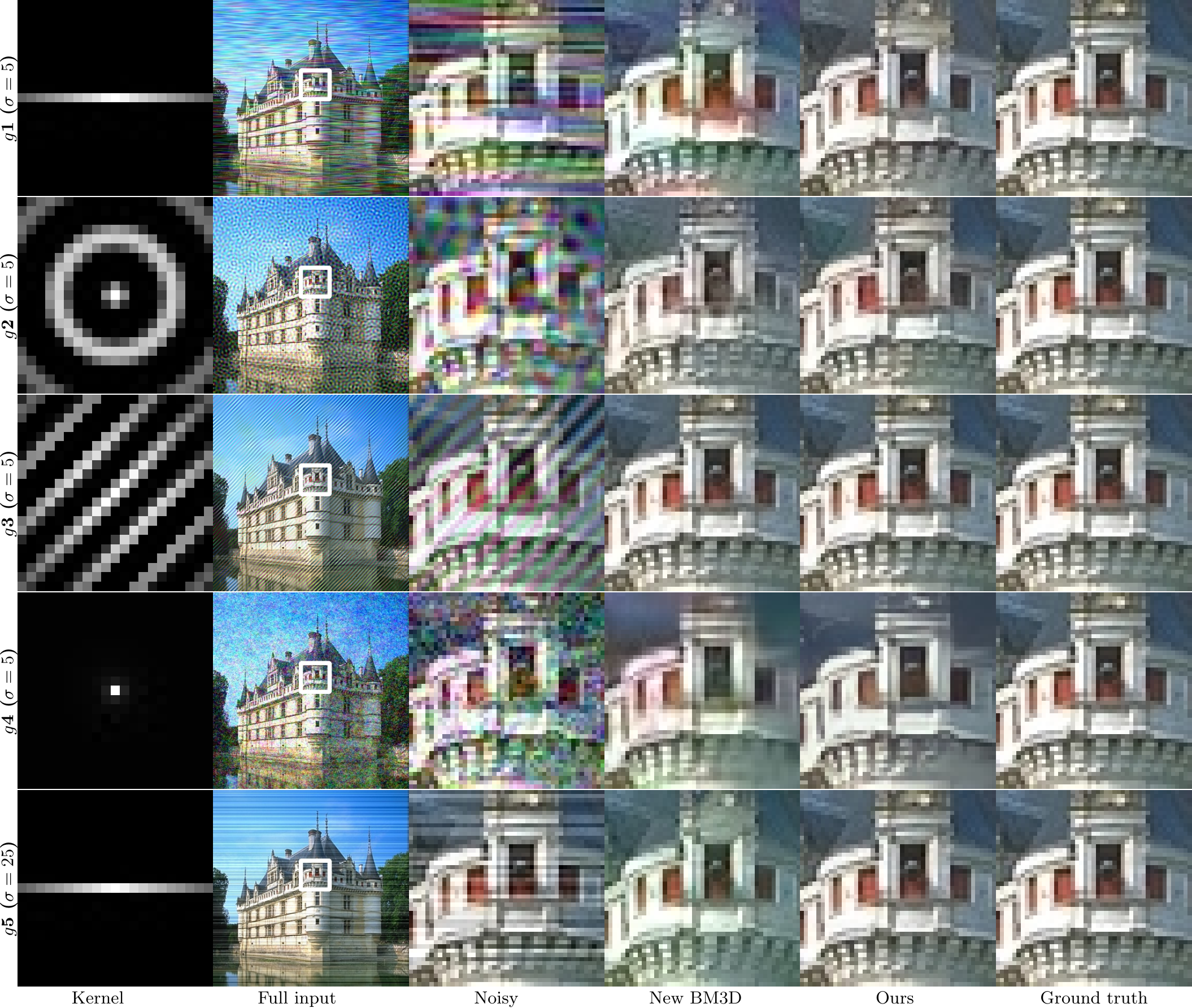}
   \end{center}
      \caption{Visualization of five correlated kernels and denoising results on different noise levels. We compare our method with the new BM3D \cite{MakinenAF19icip} designed for correlated noise.
      } 
   \label{fig:correlated_noise}
\end{figure*}

\end{document}